%% file: main.tex
\documentclass{article}

\usepackage[english]{babel}

\usepackage[letterpaper,top=2cm,bottom=2cm,left=3cm,right=3cm,marginparwidth=1.75cm]{geometry}

\usepackage{amsmath, amssymb}
\usepackage{graphicx}
\usepackage{natbib}
\usepackage{multirow}
\usepackage{booktabs} 
\usepackage[colorlinks,linkcolor=blue,filecolor=blue,citecolor=magenta,urlcolor=blue]{hyperref}
\usepackage{algorithm}
\usepackage{algorithmic}
\usepackage{soul}
\usepackage{subcaption}
\usepackage{wrapfig}
\usepackage{bbm}
\usepackage{spverbatim}
\usepackage{inconsolata}
\usepackage[affil-it]{authblk}
\usepackage{paralist}
\usepackage{diagbox}
\usepackage{makecell}
\usepackage{color, colortbl}
\usepackage{bm}
\usepackage{multirow} 
\usepackage[most]{tcolorbox}
\usepackage{censor}

\DeclareMathOperator*{\argmax}{arg\,max}
\newcommand{\cmark}{\large\color{green}\checkmark}%
\newcommand{\xmark}{\color{red}$\mathsf{X}$}%
\newcommand{\ehpc}{$\mathtt{EHPC} $ }%

\title{Efficient Prompt Compression with Evaluator Heads for Long-Context Transformer Inference}

\date{\vspace{-5ex}}

\author[1,2]{Weizhi Fei}
\author[2]{Xueyan Niu\thanks{Correspondence to: \texttt{niuxueyan3@huawei.com}.}}
\author[3]{Guoqing Xie}
\author[3]{Yingqing Liu}
\author[2]{Bo Bai}
\author[2]{Wei Han}

\affil[1]{%
    Department of Mathematical Sciences, Tsinghua University, Beijing, China}
\affil[2]{%
    Theory Lab, 2012 Labs, Huawei Technologies Co., Ltd.}
\affil[3]{%
    Architecture \& Design, ICT Products \& Solutions, Huawei Technologies Co., Ltd.}

\begin{document}
\maketitle

\begin{abstract}
Although applications involving long-context inputs are crucial for the effective utilization of large language models (LLMs), they also result in increased computational costs and reduced performance. To address this challenge, we propose an efficient, training-free prompt compression method that retains key information within compressed prompts. We identify specific attention heads in transformer-based LLMs, which we designate as \textit{evaluator heads}, that are capable of selecting tokens in long inputs that are most significant for inference. Building on this discovery, we develop $\mathtt{EHPC}$, an \textbf{E}valuator \textbf{H}ead-based \textbf{P}rompt \textbf{C}ompression method, which enables LLMs to rapidly ``skim through'' input prompts by leveraging only the first few layers with evaluator heads during the pre-filling stage, subsequently passing only the important tokens to the model for inference. \ehpc achieves state-of-the-art results across two mainstream benchmarks: prompt compression and long-context inference acceleration. Consequently, it effectively reduces the complexity and costs associated with commercial API calls. We further demonstrate that \ehpc attains competitive results compared to key-value cache-based acceleration methods, thereby highlighting its potential to enhance the efficiency of LLMs for long-context tasks.
\end{abstract}

\section{Introduction}

Large language models (LLMs) have exhibited exceptional capabilities in a variety of real-world tasks and applications, with an increasing need for processing long inputs in areas such as literary novels, legal documents, instruction manuals, and code documentation. Inference tasks that requires understanding of long contexts, such as long document summarization \citep{zhang2024benchmarking}, reasoning \citep{fei2024retrieval}, and autonomous agents \citep{singh2024llm, chen2024can}, are of particular importance due to the high stakes in these scenarios. However, the deployment of LLMs is challenged by the computational and memory demands inherent to transformer-based architectures, resulting in increased latency, particularly when processing lengthy input prompts.

\begin{wraptable}{r}{0.5\textwidth}
\caption{Overall comparison of the proposed method in terms of average performance and latency on the LongBench dataset, under the constraint of a compressed prompt length of 2048 tokens. For comprehensive results, please see Table~\ref{tab:results_prompt_compression} and Table~\ref{tab: time}.}\label{tab:table-overview}
\begin{center}
\begin{small}
\setlength{\tabcolsep}{3.5pt}
\begin{tabular}{l|ccc}
\toprule
\textbf{Method} & Performance & Latency  & \makecell{Training-free}  \\
\midrule 
$\mathtt{LongLLMLingua}$ & $48.0$ & $67.44$ &   \cmark \\
$\mathtt{LLMLingua}$ & $34.6$ & $7.51$ &  \cmark \\  
$\mathtt{LLMLingua}$-2 & $39.1$ & $1.27$ & \xmark \\ 
$\mathtt{EHPC}$ (\textit{ours}) & $\bm{49.6}$  & $\bm{0.88}$ & \cmark \\
\bottomrule
\end{tabular}
\end{small}
\end{center}
\vskip -0.3in
\end{wraptable}

Prompt compression, which entails substituting the input prompts provided to a language model with more succinct versions, has surfaced as a promising strategy for enhancing long-text understanding and mitigating associated costs.  Current mainstream methods, such as SelectContext \citep{li2023compressing}, LLMLingua~\citep{jiang2023llmlingua}  and LongLLMLingua \citep{jiang2023longllmlingua}, typically rely on pre-trained LLMs, utilizing the logits or perplexity of the prompts to evict tokens deemed insignificant. These approaches often necessitate chunking long texts for processing, leading to numerous repeated calls of the LLM and consequently incurring considerable time complexity. More efficient compression techniques, such as LLMLingua-2 \citep{pan2024llmlingua}, generally demand the training of a secondary, smaller model on labeled datasets. While these methods reduce compression time, they also incur substantial training expenses and may exhibit performance drop in out-of-distribution contexts when compared to the direct utilization of the LLM as a compressor. In this paper, we present an \textbf{E}valuator \textbf{H}ead-based \textbf{P}rompt \textbf{C}ompression method, dubbed $\mathtt{EHPC}$, which is built upon the efficient pre-filling stage of LLMs. Our method leverages the intrinsic attention mechanisms of the LLM, thus being both training-free and computationally efficient, achieving state-of-the-art (SoTA) performance across mainstream benchmarks (see Table~\ref{tab:table-overview}).

The attention mechanism is pivotal in transformer-based LLMs, aggregating information from input prompts via attention scores. In our \ehpc method, we utilize high attention scores to identify and retain significant tokens. This approach is feasible primarily because the attention scores of tokens in long texts are sparse, explained by the widely studied ``attention sink'' phenomenon \citep{xiao2023efficient, gu2024attention}. This phenomenon is characterized by LLMs' frequently assigning high attention weights to the semantically inconsequential initial token, \texttt{<BOS>}. Additionally, much of the research \citep{zhang23h2o, li2024snapkv, ge2024model, cai2024pyramidkv, wu2024retrieval, tang2024razorattention, xiao2024duoattention, fu2024lazyllm} has focused on compressing the key-value (KV) cache by eliminating entries with low attention weights. Furthermore, \ehpc can be efficient because the attention computations involved during the pre-filling stage are highly parallelizable.

\begin{figure}[t]
\centering
  \includegraphics[width=\columnwidth]{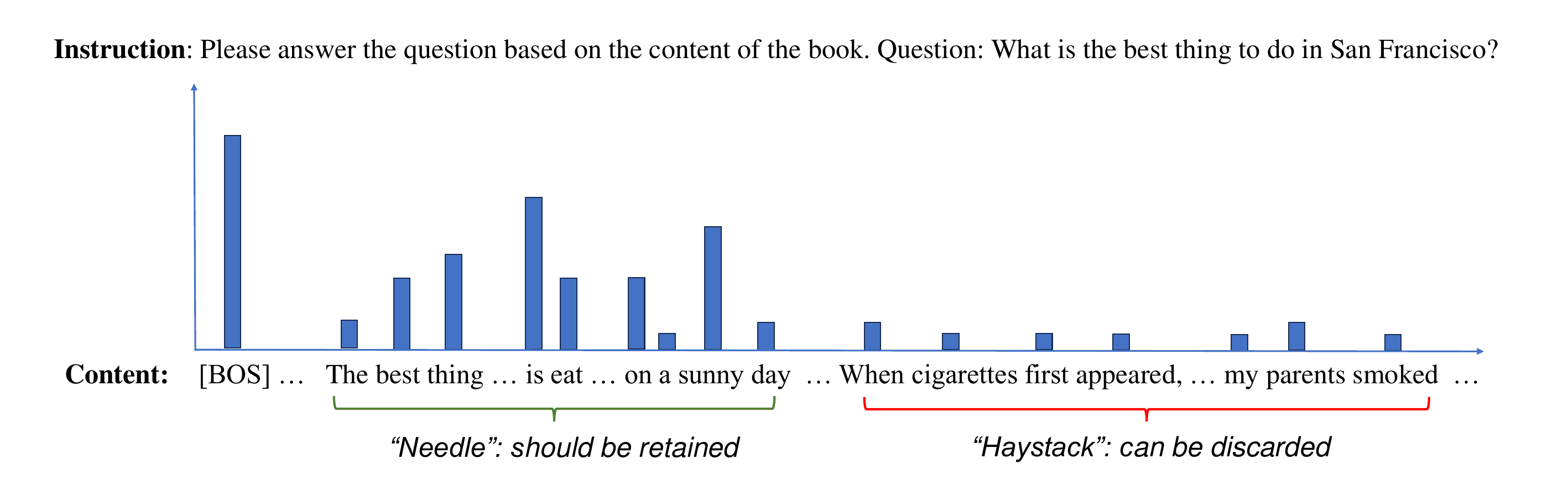}
  \caption{Visualization of attention scores from a single attention head during inference on the ``Needle-in-a-Haystack'' long-text benchmark. This benchmark requires the LLM to follow instructions and retrieve ``needles'' -- specific pieces of information randomly inserted into a long text -- to answer a given question. The evaluator heads are identified as those that accurately locate the relevant facts, thereby achieving high scores.}
  \label{fig:deletion}
\end{figure}

Unlike our prompt compression method, which retains important tokens based on the scores of evaluator heads, most KV cache compression methods preserve the cache according to all heads. Recent research \citep{wu2024retrieval, tang2024razorattention, xiao2024duoattention} has identified that certain layers and attention heads play a more significant role in processing long contexts. We have pinpointed a subset of these attention heads, which we designate as \textit{evaluator heads}, allowing LLMs to focus on essential information for inference from any position within the input sequence. To identify these evaluator heads, we design a pilot experiment using synthetic data (see Figure~\ref{fig:deletion}). We then conduct extensive experiments to demonstrate the robustness of the evaluator heads and their applicability to practical scenarios using real-world datasets. Subsequently, we applied the evaluator heads to develop a prompt compression approach $\mathtt{EHPC}$, utilizing the scores given by these heads to select tokens for inference. The proposed \ehpc requires the local development of LLMs for prompt compression and offers two application settings: Extended Model Inference (EMI) and Native Model Inference (NMI). We demonstrate its effectiveness through two important benchmarks: prompt compression and long text acceleration. When prompts compressed using \ehpc are employed in commercial models, \ehpc effectively reduces API costs while enhancing the performance of API outputs. Compared to existing prompt compression methods \citep{li2023compressing, jiang2023longllmlingua,jiang2023llmlingua}, our approach achieves new state-of-the-art (SoTA) performance and is more efficient, requiring less compression time. Moreover, when applied to native models deployed locally, prompts compressed with \ehpc accelerate long-text inference by reducing the memory usage and achieve competitive results compared to the SoTA KV cache compression methods~\citep{li2024snapkv, zhang23h2o}.
Specifically, our contributions are as follows:
\begin{itemize}
    \item We identify specific attention heads within transformer-based LLMs, which we designate as \textit{evaluator heads}, that are capable of selecting tokens in long inputs that are significant for inference.
    \item We develop $\mathtt{EHPC}$, an efficient prompt compression technique that enables LLMs to quickly ``skim through'' input prompts by utilizing only the first few layers with the evaluator heads, and then pass only the important tokens to the model for inference.
    \item We demonstrate that \ehpc has lower complexity compared to prior prompt compression method and achieves a new SoTA on the prompt compression benchmarks over LongBench and ZeroScroll, effectively reducing the API cost and memory usage of commercial LLMs.
    \item We empirically demonstrate that \ehpc is capable of accelerating long-context understanding, achieving competitive performance relative to KV cache compression methods. Notably, \ehpc improves upon direct inference by up to $40\%$ on the question-answering datasets.
\end{itemize}

\section{Related Work}
Two predominant approaches are utilized for accelerating LLMs: implicit methods that reduce the KV cache, and explicit methods that decrease the number of tokens. We briefly review these methods with a focus on the explicit methods, which can be applied to black-box LLMs such as GPT-3.5-Turbo.

\subsection{Implicit Token Reduction}\label{sec: implicit token redution}

The key-value (KV) cache reduces redundant calculations and enhances decoding efficiency by storing key and value matrices from previous tokens \citep{liu2024scissorhands,adnan2024keyformer}. However, as the input length increases, the memory requirements of the KV cache grow, creating significant challenges for long-context processing.  It was found that a small number of tokens account for the majority of the value during the computation of attention scores, leading to the proposal of the H2O~\citep{zhang23h2o} that retains only the KV cache for ``heavy hitters'', which are tokens with high attention scores. FastGen \citep{ge2024model} introduces a dual-phase adaptive KV compression strategy that includes four KV cache compression policies and dynamically evicts caches during generation based on optimal policies identified through profiling. SnapKV \citep{li2024snapkv} demonstrates that specific prompt attention patterns can be identified through an observation window at the end of prompts, and it compresses KV caches by selecting clustered attention scores via pooling operations.
\citet{wu2024retrieval} experimentally investigate how transformer-based models retrieve relevant information from arbitrary locations within long contexts, identifying certain heads, termed \textit{retrieval heads}, as crucial in this process. Subsequently, building upon the concept of \textit{retrieval heads}, several head-wise KV cache compression methods have been proposed \citep{tang2024razorattention, xiao2024duoattention}. These methods specifically preserve the KV cache of retrieval heads to maintain their functionality.

\subsection{Explicit Prompt Compression}

\paragraph{Semantic compression}
\citet{wingate-etal-2022-prompt} use soft prompts to condense context, ensuring that the compressed prompts retain a significant amount of information. \citet{Chevalier2023AdaptingLM} introduce AutoCompressor, which  adapts LLMs for compressing lengthy contexts into concise summary vectors. Similarly, research by \citet{mu2023learning} and \citet{ge2024incontext} has focused on learning gist tokens to compress context through prefix-tuning \citep{li-liang-2021-prefix}.  \citet{fei2024extending} implement a summarization model to semantically compress input text through a divide-and-conquer strategy. 

\paragraph{Token deletion}
A widely adopted approach among explicit methods is the direct removal of tokens \citep{jha2024characterizing,shi2024discovering}. Selective-Context~\citep{li2023compressing} utilize the logits of the language model to calculate the mutual information of tokens, subsequently removing tokens based on this metric. LLMLingua \citep{jiang2023llmlingua} initially computes the perplexity of each token and then integrates a budget controller with a coarse-to-fine, token-level iterative compression algorithm. Expanding on LLMLingua, LongLLMLingua~\citep{jiang2023longllmlingua} introduces the concept of conditional perplexity to intensify the focus on key information in accordance with task-specific instructions, which achieves great improvement over long text situation.  LLMLingua-2 \citep{pan2024llmlingua} represents a fast prompt compression method, as it employs a small classification model to predict the significance of each token in the prompts. This classification model takes prompt compression as a token classification task and is trained utilizing a compact transformer-based encoder on a labeled dataset.

\section{Methodology}
We present our prompt compression method, $\mathtt{EHPC}$, which is characterized by the identification and utilization of \textit{evaluator heads}. We find that in LLMs,  certain attention heads, which we designate as \textit{evaluator heads}, can be utilized to rapidly determine which tokens can be omitted during the pre-filling stage. Background information on the basic implementation of the multi-head attention mechanism in LLMs, with a focus on the enhanced efficiency of the pre-filling stage through the application of KV cache, is provided in Appendix~\ref{sec:app:prelim}.

\subsection{Prompt Compression Strategy} 
We represent the input prompt as a sequence of tokens, $\bm{x} = (x_1, x_2, \ldots, x_N)$, where $N = |\bm{x}|$ denotes the sequence length. Let $f$ be a language model. The objective of prompt compression is to identify a shorter sequence $\hat{\bm{x}}$ to replace the original sequence $\bm{x}$, which can be mathematically formulated as 
$       \min_{\hat{\bm{x}}} \mathcal{D}(f(\cdot|\bm{x}), f(\cdot|\hat{\bm{x}})),\ 
    \text{s.t.}\  |\hat{\bm{x}}| \leq |\bm{x}|,$
where $f(\cdot| \hat{\bm{x}})$ represents the conditional distribution of the language model over the input prompts, and $\mathcal{D}$ is a divergence measuring the difference between the distributions.

Analogous to the way human readers often skip words during speed reading, \ehpc employs a token deletion strategy as in  ~\citep{li2023compressing,jiang2023llmlingua} and compresses the prompt by dropping non-essential tokens directly.  In contrast to generating new context through summarization ~\citep{fei2024extending}, the token deletion strategy effectively simplifies the problem by narrowing the search space for prompt optimization. The token-level dropping strategy can be seamlessly integrated at the sentence/paragraph level ~\citep{liskavets2024prompt}.

\subsection{Evaluator Heads}

We seek to retain important tokens according to the attention scores. As detailed in Appendix~\ref{sec:multi-head attention}, each attention head conducts a weighted average over preceding tokens, wherein tokens assigned lower attention weights contribute less to the information processed by the attention heads. 
Empirical studies have demonstrated that the contribution of attention heads to the capacity for handling long contexts in LLMs is not equally important \citep{wu2024retrieval, tang2024razorattention}. Specifically, certain retrieval heads are essential and must be maintained during KV cache compression, as their removal would significantly impair the LLMs' ability to manage long contexts. We posit that specialized heads, which we designate as \textit{evaluator heads}, play a pivotal role in assessing the significance of input prompts, and these evaluator heads alone are sufficient for evaluating the tokens.

We conducted experiments based on synthetic data with known evidence to explore and verify the existence of the \textit{evaluator heads} and that they can effectively identify the crucial information in the input prompts. For a transformer model $f$ with $L$ layers and $H$ heads in each layer, we define the evaluator heads as the set $\mathcal{C}_f = \{(l_1,h_1), \cdots, (l_m,h_m)\}$, where $1 \leq l_i \leq l_j \leq L$ and $1 \leq h_i \leq h_j \leq H$ for $i \leq j$. These specialized heads are identified, and their attention scores are employed to give the final score representing the utility of each token within the input prompts. We empirical investigate properties of evaluator heads, including the existence, generalizability, and robustness, in Section~\ref{sec:EH-properties}.

The primary distinction between the \textit{evaluator heads} we have defined and the \textit{retrieval heads} discussed in Section~\ref{sec: implicit token redution} lies in their respective functions: evaluator heads are designed to assess the significance of tokens within input prompts, while retrieval heads are intended to maintain the essential KV cache. Although both types of heads aim to identify the most salient components among all attention heads through a data-driven approach, evaluator heads are deemed sufficient for their purpose, whereas retrieval heads are necessary for preserving the integrity of the KV cache.
Furthermore, evaluator heads operate explicitly on the tokens, in contrast to the \textit{retrieval heads}, which function implicitly on the KV cache.

\subsection{Pilot Experiments for Detecting Evaluator Heads}\label{sec:pilot}

We design a pilot experiment to identify the evaluator heads. By inputting a long prompt containing known evidence, we computed the corresponding scores for the evidence to pinpoint the evaluator heads. 

\begin{figure}[t]
\centering
  \includegraphics[width=0.95\columnwidth]{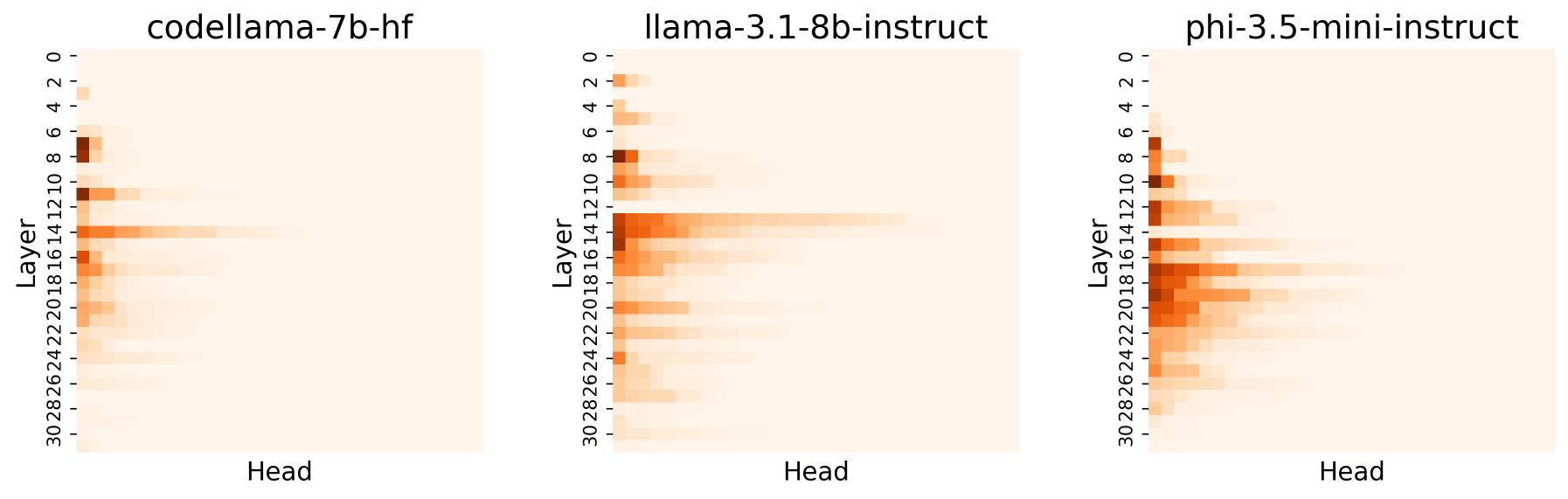}
  \caption{Heatmap of evidence scores for three different LLMs in the pilot experiment, illustrating scores across layers and heads, with heads re-ranked in descending order for clarity.}\label{fig: heatmap}
\end{figure}
Specifically, let $\hat{\bm{x}} = (x_1, x_2, \ldots, x_N)$ represent the long input sequence of $N$ tokens, and let the relevant evidence $\bm{e}$ be denoted as the sub-sequence $\bm e=(\bm{x}_{I_{\bm e}})$ of $\hat{\bm{x}}$, where $I_{\bm{e}} \subset [N]$ indicates the indices of the evidence. Let $\bm{A}^{hl} \in \mathbb{R}^{H \times L}$ denote the matrix of attention scores for layer $l$ and attention head $h$ using an LLM $f$. We then extract the last row of the attention matrix, $\bm{a}^{hl} = \bm{A}^{hl}[N, :] \in \mathbb{R}^{N}$, to represent the scores for the importance of each token, as these scores directly influence the computation of the final hidden state. To assess whether the heads are focusing on relevant information, we compute the accumulated score of the evidence as
\begin{equation}
    \hat{\bm{a}}^{hl} = \sum_{j \in I_{\bm{e}}} \bm{a}^{hl}[j].
\end{equation}
As a practical example, we utilized the ``Needle-in-a-Haystack'' benchmark \citep{NeedleInHaystack24}, a well-established long-context retrieval benchmark, to demonstrate our pilot experiments (as illustrated in Figure~\ref{fig:deletion}). Let $\hat{\bm{x}}$ be the synthesized long context, and let $\bm{e}$ represent the ``needle'' (evidence) inserted at a specific position for $f$ to identify. We recorded the accumulated score of the evidence $\hat{\bm{a}}^{hl}$ for each head. We then averaged the accumulated scores of the evidence to generate an \textit{evidence score} matrix $S \in \mathbb{R}^{H \times L}$, which we used to identify the evaluator heads. Visualizations of the evidence score matrices are presented in Figure~\ref{fig: heatmap} for several example open-source LLMs. Finally, we selected the layer with the highest score and identified the top-$k$ heads from this layer as the evaluator heads. So
\begin{align*}
    &\mathcal{C}_f = \argmax_{|\Lambda|\leq k} \  \|\bm{e}_\Lambda S\|_\mathrm{F}\\
    &\text{s.t.}\quad (i,j)\not\in \lambda,  i = \max_{1\leq l \leq L} (S\cdot \mathbf{1}_{L\times 1})_l, \forall j,
\end{align*}
where $\|\cdot\|_\mathrm{F}$ is the Frobenius norm, $\bm{e}_\Lambda = (e_{ij})$ is the incidence matrix such that $e_{ij} = 0$ if $(i,j)\not\in \Lambda$, and $e_{ij} = 1$ if $(i,j)\in \Lambda$.

\subsection{Prompt Compression Using Evaluator Heads}
Our prompt compression strategy selects salient tokens based on their averaged attention scores over the identified evaluator heads. Given a transformer-based language model $f$ and the evaluator heads $\mathcal{C}_f$ identified through the pilot experiment, for a long input prompt $\bm{x}$, we utilize the attention scores from $\mathcal{C}_f$ to compute the \textit{utility scores} $\bm s \in \mathbb{R}^{N}$ for the input during the pre-filling stage according to
\begin{equation}
    \bm s = \sum_{(l_j, h_j) \in \mathcal{C}_f} \mathbf{Pool}(\sum_{N_r \le i \le N} \frac{\bm{A}^{l_j, h_j}[i,:]}{N_o}, \  r),
\end{equation}
where $\mathbf{Pool}(\cdot, \cdot)$ denotes a pooling operation, and $r$ represents the kernel size. 
Subsequently, we employ the scores $\bm{s}$ to remove non-essential tokens, constructing the compressed prompt from the retained tokens in their original order. Although the compressed prompt retains its natural language form, it may lack certain contextual elements. To mitigate this, we adopt a pooling operation, as described in \citep{li2024snapkv}, to group neighboring tokens with similar scores, thereby generating a continuous sequence of compressed tokens that enhances readability.

While our prompt compression strategy necessitates the processing of prompts by an LLM, it leverages the computationally efficient pre-filling stage, resulting in a relatively short compression time. As depicted in Figure~\ref{fig:overview}, the proposed method reduces the number of tokens, thereby also reducing the pre-filling latency during inference.
The compressed prompts obtained are transferable and can be applied to API-based black-box large models. The application of the compressed prompt gives rise to two scenarios: Extended Model Inference (EMI) and Native Model Inference (NMI). EMI involves using a different language model to infer the compressed prompt. For instance, when applied to an API-based commercial model, this approach can reduce API latency and costs, as the API cost is linearly related to the input prompt length. NMI, in contrast, utilizes the same language model to infer the compressed prompt. In this case, our method can decrease computational memory usage and costs, akin to the KV cache compression method. In practice, the transformer-based model $f$ used for compressing prompts should exhibit robust long-context capabilities, and smaller models are preferred for efficient deployment, such as \texttt{Llama-3.1-8B}, with a context length of 128k.

\subsection{Complexity}

\begin{wrapfigure}{r}{0.5\textwidth}
    \centering
    \includegraphics[width=0.48\textwidth]{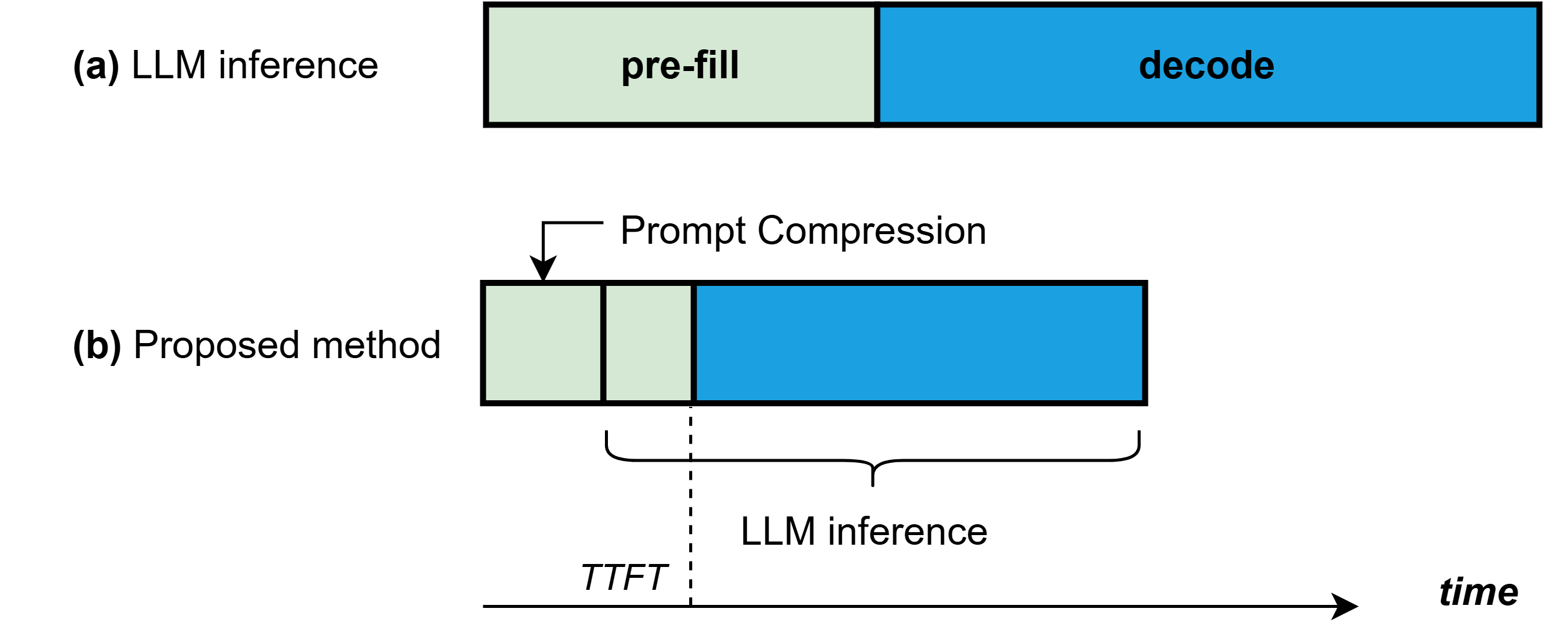}
    \caption{Illustration of the proposed method. \textbf{(a)} LLM inference comprises two stages: the pre-filling stage and the decoding stage. \textbf{(b)} The proposed prompt compression approach leverages the efficiency of the pre-filling stage, thereby reducing inference latency for both stages during inference with compressed context.}
    \label{fig:overview}
\end{wrapfigure}

We only discuss the complexity of NMI, as EMI follows a similar line of reasoning. Consider an LLM $f$ with $L$ layers, $H$ attention heads per layer, and a hidden dimension $d=d_k H$. Suppose the model $f$ is given an input prompt of $N$ tokens, and generates $t$ new tokens. In the pre-filling stage, each head computes  $\mathrm{Softmax}( \frac{\bm{Q} \bm{K}^T}{\sqrt{d_k}})$ with $\bm{Q},\bm{K}  \in \mathbb{R}^{N \times d_k}$, resulting in a complexity of $O(d_k N^2)$. Thus the total complexity for the pre-filling stage is $O(LHd_kN^2)$. During the decoding stage, the model generates $t$ tokens based on the pre-filled $\bm{K},\bm{V} \in  \mathbb{R}^{N \times d_k},$ and the total complexity is  $O(LHd_k(tN+\frac{t^2}{2}))$, with details provided in Appendix~\ref{sec:app:complexity}.

In the NMI setting, where $f$ is used for both compression and inference, our method involves two pre-filling stages and one decoding stage, as illustrated in Figure~\ref{fig:overview}. Let $\kappa_1 = L/ \max_{1\leq l \leq L} (S\cdot \mathbf{1}_{L\times 1})_l.$ The first pre-filling stage utilizes only $L / \kappa_1$ layers to compress prompts, resulting in a complexity of $O(LHd_kN^2/\kappa_1)$. Suppose the compression rate is $\kappa_2$, then the second pre-filling stage processes only $N/\kappa_2$ tokens, and the complexity is $O(\frac{LHd_kN^2}{\kappa^2_2})$. Thus the total complexity of combined pre-filling stages is $O(LHd_kN^2(\frac{1}{\kappa_1} + \frac{1}{\kappa^2_2}))$. Therefore, when $\kappa_1, \kappa_2 \geq 2,$ which is often the case, the pre-filling  with our compressing method have lower complexity since  $(\frac{1}{\kappa_1} + \frac{1}{\kappa^2_2}) \le \frac{3}{4}$. The decoding stage involves only $N/\kappa_2$ tokens, and the complexity is $O(LHd(t^2+\frac{tN}{2 \kappa_2}))$, which is naturally lower than original complexity since $\kappa_2 \geq 1$.

\section{Properties of the Evaluator Heads}\label{sec:EH-properties}
We undertake a comprehensive investigation of the evaluator heads and aim to address the following research questions:
\begin{itemize}
    \item[\textbf{(RQ1)}] \textbf{Existence:} Do evaluator heads exist significantly across LLMs? Can these heads be reliably identified through a pilot experiment utilizing synthetic data?
    \item[\textbf{(RQ2)}] \textbf{Practicality/Generalizability:} Can the evaluator heads identified through synthetic data be effectively applied to real-world long-text benchmarks?
    \item[\textbf{(RQ3)}] \textbf{Robustness:} Are evaluator heads task-oriented, i.e., do they demonstrate robustness and consistency across various downstream tasks?
\end{itemize}

To investigate (RQ1), (RQ2), and (RQ3), we conduct pilot experiments to identify evaluator heads using a synthetic benchmark in the ``Needle-in-a-Haystack'' style. Then, we evaluate these evaluator heads using real-world benchmarks, including LongBench \citep{bai2024longbench}. Details regarding the synthetic and real-world benchmarks used in our study are provided in Appendix~\ref{sec:app:data}.

\paragraph{Existence} To substantiate the existence of these heads, we conducted pilot experiments as detailed in Section~\ref{sec:pilot}, utilizing questing-answering (QA) data from the ``Needle-in-a-Haystack'' benchmark. This dataset involves the random insertion of a sentence into a variable-length context, followed by querying the LLM to retrieve the specific sentence from the long context. We assessed each head's ability to identify key information by examining the accumulated attention scores over the evidence sequence. The results, visualized in Figure~\ref{fig: heatmap}, demonstrate that several heads in the middle layers are instrumental in identifying key information relevant to the QA task within long contexts. Furthermore, the distribution of these evaluator heads is sparse and predominantly concentrated in the middle layers.

\begin{wraptable}{r}{0.5\textwidth}
\centering
\caption{Performance comparison of two sets of heads on three datasets from different tasks in the LongBench.} \label{tab:ablation of heads}
\begin{center}
\begin{small}
\begin{tabular}{l|ccc}
\toprule
   & MF-en   & Musique   & TREC \\
\midrule
Evaluator heads & 27.7 & 13.7 & 62.5 \\
\midrule
Comparison heads & 24.3 & 12.1 & 60.5 \\
\bottomrule
\end{tabular}
\end{small}
\end{center}
\end{wraptable}

\paragraph{Practicality} Next, we assess the practicality of the evaluator heads identified in our pilot experiments by evaluating their performance on downstream tasks. We select four heads with the highest scores as evaluator heads from the layer with the highest cumulative score, and we also select four additional heads from the same layer for comparison. The performances of these two sets of heads are evaluated across three downstream tasks from the LongBench benchmark. The results are presented in Table~\ref{tab:ablation of heads}. The findings indicate that the evaluator heads identified outperform the other heads on real-world data, thereby demonstrating the practical utility of the evaluator heads.

\begin{wraptable}{r}{0.5\textwidth}
\centering
\vskip -0.2in
\caption{Evaluation of task-aware approach on multi-hop reasoning and code completion from corresponding LongBench datasets. Results are averaged over the complete data. ``Evaluator heads (QA)'' refers to the heads identified using QA data from the pilot experiment, while ``Task-aware heads'' refers to heads identified from generated data customized to the respective tasks.} \label{tab:ablation of task}
\begin{center}
\begin{small}
\begin{tabular}{l|cc}
\toprule
           & Multi-hop  & Code \\
\midrule
Evaluator heads (QA)         & 16.40     & 47.86 \\
\midrule
Task-aware heads & 16.87     & 47.72 \\
\bottomrule
\end{tabular}
\end{small}
\end{center}
\vskip -0.1in
\end{wraptable}

\paragraph{Robustness} Subsequently, we assess the robustness of our approach on tasks distinct from QA in the pilot experiment. To this end, we generate probe data for two additional tasks: code completion and multi-hop variable tracking. Details of the construction of probe data is provided in Appendix~\ref{sec:app:probe}. The goal is to identify the corresponding heads for these scenarios, in addition to the simple QA data. We then evaluate three types of heads on their corresponding reasoning and coding tasks from the LongBench dataset. The results are presented in Table~\ref{tab:ablation of task}. Our findings indicate that the heads identified using customized tasks do not enhance performance on their respective downstream tasks significantly. This demonstrates that the evaluator heads are task-agnostic and that the heads identified from the simple QA data exhibit robustness in downstream applications.

\section{Experimental Results}
Having established the existence, practicality, and robustness of the evaluator heads, we now experimentally assess the effectiveness and efficiency of \ehpc in tasks aiming at reducing the API costs of commercial models and accelerating long-context inference in LLMs. We also compare the efficiency of our prompt compression method to other acceleration methods under the same memory usage in Section~\ref{sec:result-acceleration}.

\subsection{Prompt Compression Benchmark}\label{sec:result-prompt_compression}

\begin{table*}[t]
    \centering
    \caption{Performance of various prompt compression methods under different compressed length constraints on LongBench and ZeroSCROLLS. Higher values indicate better performance. The best scores are highlighted in {\bf boldfaced}.} \label{tab:results_prompt_compression}
    \vskip 0.1in
    \setlength{\tabcolsep}{1.5mm}
    \resizebox{\linewidth}{!}{
    \input{tabs/prompt_com}
    }
\end{table*}

We evaluate the prompt compression benchmark on LongBench and ZeroSCROLLS, as established by \citet{jiang2023longllmlingua}, with the objective of enhancing the quality of compressed prompts for commercial models. Prompt compression is essential for mitigating costs associated with commercial LLMs, given that API fees are often proportional to prompt length. We compare our EMI setting, which employs a local LLM for prompt compression and another model for inferring the compressed prompts, against the following baselines.

\paragraph{Baselines} We compare our method with Retrieval Augmented Generation (RAG) and other SoTA prompt compression methods. The retrieval models considered include BM25, SentenceBERT~\citep{reimers-gurevych-2019-sentence}, and OpenAI Embedding. For prompt compression methods, we evaluate against Selective-Context~\citep{li2023compressing}, LLMLingua~\citep{jiang2023llmlingua}, LongLLMLingua~\citep{jiang2023longllmlingua}, and LLMLingua-2~\citep{pan2024llmlingua} as baselines. Detailed descriptions of these prompt compression baselines are provided in Appendix~\ref{app: baselines}. LongLLMLingua extends LLMLingua by incorporating task-specific information to enhance long-context compression, while LLMLingua-2 improves efficiency through the use of a smaller model.

\paragraph{Implementation Detail} To ensure reproducibility, we employ greedy decoding and set the temperature to 0 in all experiments. For prompt compression, we utilize the  \texttt{Llama-3.1-8B} model\footnote{available at: \url{https://huggingface.co/meta-llama/Meta-Llama-3-8B}}.  Additional details of hyper-parameters are provided in Appendix~\ref{sec:app:hyper}.

\paragraph{Results}
The results of the prompt compression benchmarks are presented in Table~\ref{tab:results_prompt_compression}. Consistent with previous research~\citep{jiang2023longllmlingua}, we report the averaged results for ZeroSCROLLS and LongBench, as well as the average performance for the sub-tasks on LongBench, with target compressed prompt lengths of $2000$ and $3000$ tokens. Table~\ref{tab:results_prompt_compression} illustrates that our method achieves superior performance on average across both benchmarks under the length constraints. The results show that \ehpc surpasses the original prompts and significantly enhances the accuracies of the QA tasks, indicating its effectiveness in retrieval tasks by alleviating the disturbance of noisy context. Notably, our model performs well on code tasks in the LongBench, which can be attributed to the identification of the key tokens involved in the inference of coding tasks. Additionally, the performance of \ehpc on summarization and few-shot learning tasks is competitive. Overall, our compression method achieves high-quality compression, even outperforming the original prompts on specific tasks.

\paragraph{Compression Latency}
\begin{table}[t]
\caption{The averaged time (in seconds) for different methods applied to a subset of LongBench, targeting compression to 2,048 tokens. The subset contains 10 examples from RepoBench-P, with an average of 14,354 tokens. Each example was repeated 5 times to reduce randomness.  Specifically, \texttt{ChatGPT-3.5-Turbo} (all tokens) serves as the baseline using the original prompts, where the inference time is carried over from ~\citep{liskavets2024prompt}.} \label{tab: time}
\vskip 0.15in
\begin{center}
\begin{small}
\begin{tabular}{l|ccc}
\toprule
\diagbox{Method}{Latency} & \makecell{Compression\\ time}  & \makecell{Inference time\\ (2048 tokens)} & \textbf{Total} \\
\midrule
LLMLingua-2 & $7.51$& $\mathbf{1.09}$       & $8.60$   \\  
LongLLMLingua & $67.44$ & $1.31$       & $68.74$   \\
LLMLingua-2 & $1.27$ & $1.15$       & $2.37$   \\  
\ehpc (\textit{ours}) & $\mathbf{0.88}$ & $1.11$       & $\mathbf{1.99}$    \\ 
\midrule
\multicolumn{3}{c}{\texttt{ChatGPT-3.5-Turbo} (all tokens)} &  $2.16$ \\
\bottomrule
\end{tabular}
\end{small}
\end{center}
\end{table}

We evaluate the running times of various methods using a subset of LongBench, including direct inference, LLMLingua-2, LongLLMLingua, and our proposed approach. The results are presented in Table~\ref{tab: time}. Each prompt compression method reduces the average prompt length from 14,354 tokens to 2,048 tokens. The experiments were conducted on a GPU with 40GB of VRAM. For a fair comparison, we utilized the same language model, \texttt{Llama-3.1-8B}, for all methods except LLMLingua-2, which requires a specialized model. The results in Table~\ref{tab: time} indicate that our method is significantly faster than LongLLMLingua and also outperforms LLMLingua-2, which is known for its efficiency. The lower latency of our compression strategy is attributed to its reliance on the efficient pre-filling stage.

\subsection{Acceleration of LLM Inference}\label{sec:result-acceleration}

\begin{table*}[t]
\caption{Performance of different acceleration methods over various LLMs under the same KV cache memory usage. Higher values indicate better performance. The best scores are {\bf boldfaced}.
}\label{tab:kv longbench}\vskip 0.1in
    \centering 
\resizebox{\textwidth}{!}{\input{tabs/kv_cache}}
\end{table*}

We demonstrate that \ehpc effectively reduces memory overhead and computation costs during long-context inference of LLMs by compressing input prompts. By inferring over the $k$ times compressed prompt, the KV cache during pre-filling is reduced $k$ times, also accelerating the decoding stage. In the NMI setting, where the same model is used for both compression and inference, the achieved acceleration is comparable to that of KV cache compression methods with the same compression ratio. Therefore, we compare our method against the KV cache compression method on the long-text acceleration benchmark of LongBench.

\paragraph{Baselines}
We consider acceleration frameworks that include KV cache compression and prompt compression as baselines. For KV cache compression methods, we select SoTA approaches, including  H2O~\citep{zhang23h2o} and SnapKV~\citep{li2024snapkv}, which eliminate unnecessary KV caches based on attention scores. For the prompt compression method, we select GemFilter~\citep{shi2024discovering}, which utilizes attention scores from early layers to compress prompts.

\paragraph{Implementation Detail}
We utilize two popular long-context models: \texttt{Llama-3.1-8B} and \texttt{Phi-3.8B}, both support a context window length of 128k. For each method, we set the target lengths to 1024 and 2048 tokens for prompt compression and KV cache compression respectively.

\paragraph{Results}
In Table~\ref{tab:kv longbench}, we present the results of long-text inference acceleration under KV cache constraints of $1024$ and $2048$ tokens using various methods. \ehpc achieves the best average performance compared to the baselines across different compression rates. Relative to KV cache-based methods, our prompt compression method performs particularly well on specific tasks such as QA, even outperforming the results obtained using the full KV cache. This improvement is particularly significant when applying our method to the \texttt{Phi-3.8B} model, enhancing its performance on direct inference by an average of up to $40\%$ averaged on QA tasks, including Single-Document QA and Multi-Document QA. However, prompt compression exhibits a decline in performance on code-related tasks and few-shot learning tasks. While the effectiveness of the KV cache compression method decreases gradually as the KV cache memory is reduced, it remains more advantageous than prompt compression for these tasks. Overall, \ehpc demonstrates competitive results compared to the KV cache compression method, particularly in QA tasks.

\section{Conclusion}
In this paper, we introduce $\mathtt{EHPC}$, an efficient and effective prompt compression method that utilizes evaluator heads to leverage attention scores from the pre-filling stage of transformer-based large language models. Unlike previous speedup methods that rely on training specialized small models, our approach is training-free and achieves state-of-the-art performance in prompt compression. Specifically, we empirically identify certain attention heads, designated as evaluator heads, that effectively identify important tokens in long inputs. We investigate their existence, practicality, and robustness using both synthetic and real-world data. Subsequently, we implement our evaluator head-based prompt compression ($\mathtt{EHPC}$) method in two settings: native model inference and extended model inference. We demonstrate the effectiveness of \ehpc in these settings across two mainstream benchmarks, highlighting its potential to significantly reduce API costs for commercial use and accelerate long-text inference.

\bibliography{ref}
\bibliographystyle{natbib_style}

\newpage
\appendix
\onecolumn
\section{Background}\label{sec:app:prelim}

\paragraph{Multi-head Attention}~\label{sec:multi-head attention}
Transformer-based models autoregressively predict the next token based on the $\tau$ preceding tokens according to
\begin{align*}
    & \bm{h}^l = \bm{h}^{l-1} + \bm{\delta}^{l} + \bm{h}^l,\  m^l = \mathrm{FFN}(\bm{h}^{l-1} + \bm{\delta}^{l}) 
\end{align*}
where $\bm{h}^l, \bm{m}^l, \bm{\delta}^{l} \in \mathbb{R}^{d}$ are the hidden states of the $l$-th layer and $\mathrm{FFN}(\cdot)$ denotes the feed forward network. Readers who are interested in more details and mathematical implications can refer to \citep{niu2024beyond}.
Transformer models typically utilize multi-head attention, so that the hidden state $\bm{\delta}^{l}$ at layer $l$ is computed as
\begin{equation*}
    \bm{\delta}^{l} = \bm{W}^l_a ~ \mathrm{ConCat}( \hat{\bm{h}}^{l1}, \cdots, \hat{\bm{h}}^{lH}),
\end{equation*}
where $H$ is the number of heads in each layer, and $\hat{h}^{lh}$ denotes the hidden representation of the $h$-th head at layer $l$.
The attention operation in each attention head is
\begin{equation}\label{eq: attention}
            \hat{\bm{h}}^{lh} = \mathrm{Softmax}( \frac{\bm{Q}^{lh} (\bm{K}^{lh})^T}{\sqrt{d_k}}) \cdot \bm{V}^{lh},
\end{equation}
where $\bm{Q}^{lh}, \bm{K}^{lh}, \bm{V}^{lh} \in \mathbb{R}^{\tau \times d_k}$, and $d_k = d/H $ is the dimension of each head. 
As shown in Eq. \eqref{eq: attention}, the attention mechanism aggregates information and selects important tokens from input prompts. The weights calculated by the dot product between queries and keys determine which tokens of the values are considered within this attention block. The independent multi-head attention mechanism enables the model to capture information from previous tokens in multiple ways.

\paragraph{LLM Inference and KV Cache Compression}

Contemporary decoder-only LLMs generating new tokens with a series of tokens as input  involves two stages: the pre-filling stage and the decoding stage. During the pre-filling stage, the model processes compute the intermediate states (keys and values in attention operations), which is highly parallelized and  computational efficient. In the decoding stage, LLMs load the precomputed KV cache and generate each output token autoregressively through a forward pass,  which is slower due to its serial nature  \citep{zhong2024distserve}.

When processing long text inputs, the size of the corresponding KV cache increases dramatically, significantly raising both computational costs and time. To tackle this issue, KV cache compressing methods that eliminate the unnecessary KV caches have been proposed. Many KV cache compression methods are based on the attention weights to propose a policy to determine which tokens to retain in memory. Let $\bm{A}^{lh} = \mathrm{Softmax}( \frac{\bm{Q}^{lh} (\bm{K}^{lh})^T}{\sqrt{d_k}})$ represent the attention matrix. The policy first averages the last rows to represent the scores of input tokens, as follows:
\begin{equation}
    \bm{s} = \sum_{N_r \le i \le N} \frac{\bm{A}[i,:]}{N_o},
\end{equation}
where  $N$ denote the number of tokens in the prompt, $N_o$ is the observed windows length and $N_r$ is the length of remaining part, such that $N = N_r + N_o.$
To further improve the contextual integrity, \citet{li2024snapkv} apply the  pooling operation to perform clustering:
\begin{equation}
     \hat{\bm{s}} = \mathbf{Pool}(\bm{s}, k),
\end{equation}
where $\mathbf{Pool}(\cdot , \cdot)$ represents a pooling operation such as $\max (\cdot)$ and $\mathrm{Average}(\cdot)$, and $k$ is the kernel size. This trick ensures that the identified tokens are continuous rather than isolated, resulting in more coherent semantics

\section{Hyper-parameters}\label{sec:app:hyper}
In this section, we introduce the hyper-parameter for reproduce. We first introduce our detected  evaluator heads and then introduce the hyper-parameter to use these heads.

\paragraph{Detected evaluator heads}
We conducted pilot experiments using the "Needle-in-a-Haystack" benchmark across three popular LLMs: \texttt{Llama-3.1-8B-Instruct}, \texttt{CodeLlama-7B}, and \texttt{Phi-3.5-mini-3.8B-Instruct}.
\begin{itemize}
    \item For \texttt{Llama-3.1-8B-Instruct}, which has 32 layers and 32 heads, the selected layer is 13, and the chosen heads are $[18,13,21,8,11,1,4,3]$.
    \item For \texttt{CodeLlama-7B}, also with 32 layers and 32 heads, the selected layer is 14, and the selected heads are $[24,3,18,7,29,2,9,1]$.
    \item Finally, for \texttt{Phi-3.5-mini-3.8B-Instruct}, which features 32 layers and 32 heads, the selected layer is 17, and the chosen heads are $[7,17,30,2,6,16,25,18]$.
\end{itemize}

\paragraph{Using evaluator heads}
The hyperparameters applied during the evaluation of heads include the size of the observed windows, the pooling operation, and the kernel size for pooling. In all experiments, we used the average pooling operation, as the difference between average pooling and maximum pooling was negligible experimentally.
For \texttt{Llama-3.1-8B-Instruct}, we set the size of the observed windows and the kernel size for pooling to 16 and 32, respectively. For \texttt{Phi-3.5-mini-3.8B-Instruct}, the size of the observed windows and the kernel size for pooling were set to 4 and 32, respectively. A larger kernel size typically results in a more continuous compressed context, which is why we generally prefer using a larger kernel size.

\section{Complexity}\label{sec:app:complexity}
We further discuss the complexity of NMI in this section. Consider an LLM $f$ with $L$ layers, $H$ attention heads per layer, and a hidden dimension $d=d_k H$. Suppose the model $f$ is given an input prompt of $N$ tokens, and generates $t$ new tokens. In the pre-filling stage, each head computes  $\mathrm{Softmax}( \frac{\bm{Q} \bm{K}^T}{\sqrt{d_k}})$ with $\bm{Q},\bm{K}  \in \mathbb{R}^{N \times d_k}$, resulting in a complexity of $O(d_k N^2)$. Thus the total complexity for the pre-filling stage is $O(LHd_kN^2)$. During the decoding stage, the model generates $t$ tokens based on the pre-cached $\bm{K},\bm{V} \in  \mathbb{R}^{N \times d_k}.$ At the $i+1$ tokens generated, the softmax operation in each head involved $\bm{Q}\in \mathbb{R}^{1 \times d_k},\bm{V} \in  \mathbb{R}^{(N+i)\times d_k}$ and the complexity is $O(d_k (N+i))$. to calculate the total complexity for generating 
$t$ tokens, we sum the complexities across all $t$ tokens: 
\begin{equation}
    O(\sum_{i=1}^{t-1} d_k (N+i)) = O( d_k (Nt + \frac{(t-1)^2}{2})) = O( d_k (Nt + \frac{t^2}{2})).
\end{equation}
Thus, the overall complexity for generating $t$ tokens during the decoding stage is $O(LHd_k (Nt+t^2))$.

Regarding the complexity of EMI, it remains similar if the inference model is still transformer-based.

\section{Datasets}\label{sec:app:data}
We present the synthetic and real-world benchmarks for assessing the capability of LLMs in handling long contexts.
\subsection{Synthetic Benchmarks}

\paragraph{Needle-in-a-Haystack}~\citep{NeedleInHaystack24} is a well-known synthetic dataset used to benchmark the long context ability. It involves randomly inserting a sentence into a variable-length long context, followed by querying a given LLM to retrieve that specific sentence from long context.

\paragraph{Ruler}~\citep{hsieh2024ruler} is a synthetic long-context benchmark that extends ``Needle-in-a-Haystack'' by offering more complex tasks. RULER encompasses diverse types and quantities of needles and introduces new task categories, such as multi-hop tracing and aggregation, to evaluate behaviors beyond direct searching within the context. 

\subsection{Real-world Benchmarks}

\textbf{LongBench}~\citep{bai2024longbench} is a widely used long-context benchmark that includes 21 datasets across six types of tasks. The six tasks are single-document question answering, multi-document question answering, summarization, few-shot learning, code completion, and synthetic tasks for retrieval and counting. In line with previous research~\citep{jiang2023llmlingua, jiang2023longllmlingua}, we focus on English datasets and encompass six tasks across 16 datasets. We present the detailed introduction in the following context.

\paragraph{Single-Doc QA}
\begin{itemize}
\item \textbf{NarrativeQA} \citep{kovcisky2018narrativeqa} is a standard question-answering dataset that includes texts from Project Gutenberg and movie screenplays sourced from various websites. It contains 200 entries and is evaluated using the F1 metric.
\item \textbf{Qasper} \citep{dasigi2021dataset} is a question-answering dataset focused on NLP publications, featuring abstractive, extractive, and yes/no questions. It consists of 200 entries and is evaluated using the F1 metric.
\item \textbf{MultiFieldQA-en}~\citep{bai2024longbench} is created from diverse sources, including legal documents, government reports, encyclopedias, and academic publications. It includes 150 entries and is evaluated using the F1 metric.
\end{itemize}
\paragraph{Multi-Doc QA}
\begin{itemize}
\item \textbf{HotpotQA} \citep{yang2018hotpotqa} features many 2-hop questions crafted by native speakers, based on two related paragraphs. It contains 200 entries and is evaluated using the F1 metric.
\item \textbf{2WikiMultihopQA} \citep{ho-etal-2020-constructing} includes up to 5-hop questions systematically constructed with manual templates. Answering these questions requires reasoning paths that cannot be resolved by local content alone. It contains 200 entries and is evaluated using the F1 metric.
\item \textbf{MuSiQue}~\citep{trivedi-etal-2022-musique} consists of up to 4-hop questions, eliminating shortcuts and questions about naturalness. Each question includes 2-4 supplementary paragraphs that outline the reasoning path and relevant content. It comprises 200 entries and is evaluated using the F1 metric.
\end{itemize}
\paragraph{Summarization}
\begin{itemize}
\item \textbf{GovReport} \citep{huang-etal-2021-efficient} collects comprehensive reports containing human-written summaries from the U.S. Government Accountability Office and Congressional Research Service, covering a wide array of national policy issues. It includes 200 entries and is evaluated using the Rouge-L metric.
\item \textbf{QMSum} \citep{zhong2021qmsum} contains annotated pairs of meeting summaries across various domains, including product, academic, and committee meetings. It includes 200 entries and is evaluated using the Rouge-L metric.
\item \textbf{MultiNews}~\citep{fabbri2019multi} is a multi-document summarization dataset that clusters 2-10 news articles discussing the same event or topic, each paired with a human-written summary, thus forming a new long-text summarization task. It includes 200 entries and is evaluated using the Rouge-L metric.
\end{itemize}
\paragraph{Few-Shot Learning}
To construct few-shot learning with long text, \citep{bai2024longbench} selected a range of training examples from the following datasets to concatenate the context in LongBench:
\begin{itemize}
\item \textbf{TREC} \citep{li2002trec} is a classification dataset featuring fine-grained class labels. It includes 200 entries and is evaluated using the accuracy metric.
\item \textbf{TriviaQA} \citep{zhong2021qmsum} is another classification dataset that involves messenger-like conversations accompanied by human-written summaries. It contains 200 entries and is evaluated using the F1 metric.
\item \textbf{SAMSum}~\citep{fabbri2019multi} is a reading comprehension dataset consisting of question-answer pairs annotated with evidence passages. It includes 200 entries and is evaluated using the Rouge-L metric.
\end{itemize}

\paragraph{Code Completion}
Code completion is a critical yet challenging task utilized by auto-completion systems to assist users in predicting and completing code based on previous inputs and context.
\begin{itemize}
\item \textbf{LCC} \citep{Guo2023LongCoderAL}: Sampled from the Long Code Completion dataset, this dataset is constructed by filtering code based on length within individual GitHub files. It incorporates preceding lines of code as context, with the next line serving as the answer. This dataset includes 200 entries and is evaluated using the Exact Match (EM) metric.
\item \textbf{RepoBench-P} \citep{liu2023repobench}: Collected from GitHub repositories, this dataset aggregates relevant cross-file code snippets based on module import statements. These snippets are combined with preceding lines of code in the current file to predict the next line of code, utilizing the most challenging XF-F setting. It comprises 200 entries and is evaluated using the Exact Match (EM) metric.
\end{itemize}
\paragraph{Synthetic Task}
Two synthetic datasets evaluate the ability to retrieve and count from long contexts.
\begin{itemize}
\item \textbf{PassageRetrieval-en}: Derived from English Wikipedia, this dataset randomly samples 30 passages and selects one for summarization, with the task of identifying the original paragraph corresponding to the summary. It includes 500 entries and is evaluated using the Edit Similarity metric.
\item \textbf{PassageCount}: This dataset presents a more complex challenge by randomly selecting paragraphs from English Wikipedia, repeating and shuffling them. The model is required to determine the number of unique passages among the provided set. It consists of 500 entries and is evaluated using the Edit Similarity metric.
\end{itemize}

\textbf{ZeroSCROLLS} \citep{shaham-etal-2023-zeroscrolls} is a well-known long-context benchmark that encompasses three types of tasks: summarization, question answering, and aggregation across ten datasets. In line with prior research \citep{jiang2023llmlingua, jiang2023longllmlingua}, we focus on the validation set for evaluation, as it is the only set providing ground truth data. Below, we introduce the ten datasets across the three tasks, using the same evaluation metric for each dataset.

\paragraph{Summarization}
\begin{itemize}
\item \textbf{GovReport}: Contains long reports from the Congressional Research Service and U.S. Government Accountability Office, paired with expert-written summaries.
\item \textbf{SummScreenFD}: Comprises episode scripts from TV shows with community-contributed recaps sourced from Wikipedia and TVMaze.
\item \textbf{QMSum}: A query-based summarization dataset featuring meeting transcripts, including academic, industrial, and parliamentary discussions, with each instance accompanied by a specific query.
\item \textbf{SQuALITY}: A question-focused dataset derived from Project Gutenberg stories, requiring summaries based on crowdsourced guiding questions.
\end{itemize}

\paragraph{Question Answering}
\begin{itemize}
\item \textbf{Qasper}: Contains NLP papers from the Semantic Scholar Open Research Corpus, with questions based on abstracts answered by practitioners.
\item \textbf{NarrativeQA}: Features questions and answers derived from books and movie scripts, with questions crafted from summaries provided by annotators.
\item \textbf{QuALITY}: Comprises stories and articles requiring multiple-choice questions that necessitate reading substantial portions for accurate answers.
\item \textbf{MuSiQue}: Focuses on multi-hop questions using Wikipedia paragraphs, including both answerable and unanswerable questions.
\end{itemize}
\paragraph{Aggregation}
\begin{itemize} 
\item \textbf{SpaceDigest}: A sentiment aggregation task using 50 hotel reviews per hotel from the Space dataset, focusing on strictly positive or negative reviews.
\item \textbf{BookSumSort}: A task based on the BookSum dataset, requiring the reordering of shuffled chapter summaries from selected books to their original order.
\end{itemize}

\section{Baselines of prompt compression}\label{app: baselines}
\paragraph{Selective-Context} applies the logits of a causal language model to compute the self-information for each token, subsequently eliminating unnecessary content based on this self-information. 
\paragraph{LLMLingua} divides the target prompt into several segments and allocates different compression budgets according to the perplexity distribution of these segments. 
\paragraph{LongLLMLingua} further incorporates task information (such as questions for document QA) and employs a Question-Aware strategy to enhance the density of key information in long contexts. 
\paragraph{LLMLingua-2} is based on fine-tuned smaller BERT models to  improve efficiency, leveraging global information from an extractive text compression dataset annotated by ChatGPT.

\section{Additional Details for Evaluator Heads}~\label{sec:app:probe}

In this section, we describe how we construct task-aware synthetic data to identify task-aware heads. We focus on two types of tasks: multi-hop reasoning and code completion.
For multi-hop reasoning, we utilize the multi-hop tracing task from the Ruler, which is an extension of ``Needle-in-a-Haystack''. Multi-hop tracing involves randomly inserting several interconnected chains to assess how effectively the model tracks all the content of these chains in response to given questions. This forms the basis of a multi-hop reasoning task.
Regarding the code completion task, we first created a long code dataset that is unrelated to LongBench to prevent data leakage. We then manually inserted the code needing completion into the context, thereby effectively generating the necessary evidence.
These two types of data are designed to identify specific heads within their respective domains using synthetic data tailored to those domains.

\section{Example Prompt Compression Results Using \ehpc}

\begin{minipage}[c]{.5\textwidth}
    \begin{tcolorbox}[colback=blue!5!white,colframe=blue!50!black!50!,left=2pt,right=2pt,top=0pt,bottom=1pt,colbacktitle=blue!25!white,title=\textbf{\color{black} Original Prompt}]
        They took hold of each other's hands and wandered out of the big Palace. They talked about grandmother, and about the roses upon the roof. Wherever they went the winds lay still and the sun broke through the clouds. When they reached the bush with the red berries they found the reindeer waiting for them, and he had brought another young reindeer with him, whose udders were full. The children drank her warm milk and kissed her on the mouth. Then they carried Kay and Gerda, first to the Finn woman, in whose heated hut they warmed themselves and received directions about the homeward journey. Then they went on to the Lapp woman; she had made new clothes for them and prepared her sledge. Both the reindeer ran by their side, to the boundaries of the country; here the first green buds appeared, and they said 'Good-bye' to the reindeer and the Lapp woman. They heard the first little birds twittering and saw the buds in the forest. Out of it came riding a young girl on a beautiful horse, which Gerda knew, for it had drawn the golden chariot. She had a scarlet cap on her head and pistols in her belt; it was the little robber girl, who was tired of being at home. She was riding northwards to see how she liked it before she tried some other part of the world. She knew them again, and Gerda recognised her with delight.
    \end{tcolorbox}
\end{minipage}
\begin{minipage}[c]{.5\textwidth}
    \begin{tcolorbox}[colback=teal!5!white,colframe=teal!50!black!50!,left=2pt,right=2pt,top=0pt,bottom=1pt,colbacktitle=teal!25!white,title=\textbf{\color{black} Compressed Prompt}]
    They took hold \blackout{of each other's hands and wandered out of the big Palace. They talked about grandmother, and about the roses upon the roof. Wherever they went the winds lay still and the sun broke through the clouds. When they reached the bush with the red berries they found the reindeer waiting for them, and he had brought another young reindeer with him, whose udders were full. The children drank her} warm milk \blackout{and kissed her on the mouth. Then} they carried Kay and Gerda, first to \blackout{the} Finn woman. \blackout{in whose heated hut they warmed themselves and received directions about the} homeward journey \blackout{Then they went on to the Lapp woman; she had made new clothes for them and prepared her sledge. Both the reindeer ran by their side, to the boundaries of the country; here the first green buds appeared, and they said 'Good-bye' to the reindeer and the Lapp woman. They heard the first little birds twittering and saw the buds in the forest. Out of it came} riding a young girl on a beautiful horse, which Gerda knew, for it had drawn the golden chariot. She had a scarlet cap on her head and pistols in her belt; it was the little robber girl, who was tired of being at home. She was riding northwards to see how she liked it before \blackout{she tried some other part} of the world. She knew them again, and Gerda recognised her with delight.
    \end{tcolorbox}
\end{minipage}

\end{document}

%% file: tabs/prompt_com.tex
\begin{tabular}{l|ccccccc|cc|ccc}
    \toprule
        \multirow{2}{*}{Methods} &  \multicolumn{9}{@{}c}{{\bf LongBench}} &  \multicolumn{3}{@{}c}{{\bf ZeroSCROLLS}} \\
        & SingleDoc & MultiDoc & Summ. & FewShot & Synth. & Code & Avg. & \# Tokens & $\kappa_2$ & Avg. & \# Tokens & $\kappa_2$\\
    \midrule
        Original Prompt & 39.7 & 38.7 & 26.5 & 67.0 & 37.8 & 54.2 & 44.0 & 10,295 & - & 32.5 & 9,788 & -  \\
    \cmidrule (r){1-1}\cmidrule (lr){2-10} \cmidrule (lr){11-13}
    Zero-shot & 15.6 & 31.3 & 15.6 & 40.7 & 1.6 & 36.2 & 23.5 & 214 & 48$\times$ & 10.8 & 32 & 306$\times$ \\
    \midrule
    \multicolumn{6}{@{}r}{{ \textit{3,000 tokens constraint}}} \\
    \midrule
      \multicolumn{13}{@{}l}{{ \textit{Retrieval-based Methods}}} \\ 
    BM25 & 32.3 & 34.3 & 25.3 & 57.9 & 45.1 & 48.9 & 40.6 & 3,417 & 3$\times$ & 19.8 & 3,379 & 3$\times$ \\
    SBERT & 35.3 & 37.4 & 26.7 & 63.4 & 51.0 & 34.5 & 41.4 & 3,399 & 3$\times$ & 24.0 & 3,340 & 3$\times$\\
    OpenAI & 34.5 & 38.6 & 26.8 & 63.4 & 49.6 & 37.6 & 41.7 & 3,421 & 3$\times$ & 22.4 & 3,362 & 3$\times$  \\
    \cmidrule (r){1-1}\cmidrule (lr){2-10} \cmidrule (lr){11-13}
    \multicolumn{7}{@{}l}{{ \textit{Compression-based Methods}}} \\
    Selective-Context & 23.3 & 39.2 & 25.0 & 23.8 & 27.5 & 53.1 & 32.0 & 3,328 & 3$\times$ & 20.7 & 3,460 & 3$\times$ \\
    LLMLingua & 31.8 & 37.5 & 26.2 & 67.2 & 8.3 & 53.2 & 37.4 & 3,421 & 3$\times$ & 30.7 & 3,366 & 3$\times$ \\
    {LLMLingua-2} &  35.5 & 38.7 & 26.3 & 69.6 & 21.4 & 62.8 & 42.2 & 3,392 & 3$\times$ & 33.5 & 3,206 & 3$\times$ \\
    {LongLLMLingua} &  40.7 & 46.2 & \textbf{27.2} & \textbf{70.6} & \textbf{53.0} & {55.2} & 48.8 & 3,283 & 3$\times$ & 32.8 & 3,412 & 3$\times$ \\
    
    \cmidrule (r){1-1}\cmidrule (lr){2-10} \cmidrule (lr){11-13}
    
    {\cellcolor[rgb]{0.925,0.957,1}}\texttt{EHPC (EMI)} &    {\cellcolor[rgb]{0.925,0.957,1}}\textbf{44.2} & {\cellcolor[rgb]{0.925,0.957,1}}\textbf{49.1} & {\cellcolor[rgb]{0.925,0.957,1}}25.1 & {\cellcolor[rgb]{0.925,0.957,1}}68.8 & {\cellcolor[rgb]{0.925,0.957,1}}54.0 & {\cellcolor[rgb]{0.925,0.957,1}}\textbf{63.0} & {\cellcolor[rgb]{0.925,0.957,1}}\textbf{49.7} & {\cellcolor[rgb]{0.925,0.957,1}}2,892 & {\cellcolor[rgb]{0.925,0.957,1}}3$\times$ & {\cellcolor[rgb]{0.925,0.957,1}}\textbf{36.7} & {\cellcolor[rgb]{0.925,0.957,1}}3,005 & {\cellcolor[rgb]{0.925,0.957,1}}3$\times$\\
    \midrule
    \multicolumn{6}{@{}r}{{ \textit{2,000 tokens constraint}}} \\
    \midrule
      \multicolumn{13}{@{}l}{{ \textit{Retrieval-based Methods}}} \\ 
    BM25 & 30.1 & 29.4 & 21.2 & 19.5 & 12.4 & 29.1 & 23.6 & 1,985 & 5$\times$ & 20.1 & 1,799 & 5$\times$\\
    SBERT & 33.8 & 35.9 & 25.9 & 23.5 & 18.0 & 17.8 & 25.8 & 1,947 & 5$\times$ & 20.5 & 1,773 & 6$\times$ \\
    OpenAI & 34.3 & 36.3 & 24.7 & 32.4 & 26.3 & 24.8 & 29.8 & 1,991 & 5$\times$ & 20.6 & 1,784 & 5$\times$ \\
    \cmidrule (r){1-1}\cmidrule (lr){2-10} \cmidrule (lr){11-13}
    \multicolumn{7}{@{}l}{{ \textit{Compression-based Methods}}} \\
    Selective-Context & 16.2 & 34.8 & 24.4 & 15.7 & 8.4 & 49.2 & 24.8 & 1,925 & 5$\times$ & 19.4 & 1,865 & 5$\times$ \\
    LLMLingua & 22.4 & 32.1 & 24.5 & 61.2 & 10.4 & {56.8} & 34.6 & 1,950 & 5$\times$  & 27.2 & 1,862 & 5$\times$ \\
    LLMLingua-2 & 29.8 & 33.1 & 25.3 & 66.4 & 21.3 & 58.9 & 39.1 & 1,954 & 5$\times$ & 33.4 & 1,898 & 5$\times$ \\
    {LongLLMLingua} & 39.0 & 42.2 & \textbf{27.4} & 69.3 & \textbf{53.8} & 56.6 & 48.0 & 1,809 & 6$\times$ & 32.5 & 1,753 & 6$\times$ \\ 
    
    \cmidrule (r){1-1}\cmidrule (lr){2-10} \cmidrule (lr){11-13}
    
    {\cellcolor[rgb]{0.925,0.957,1}}\texttt{EHPC (EMI)} & {\cellcolor[rgb]{0.925,0.957,1}}\textbf{44.5} & {\cellcolor[rgb]{0.925,0.957,1}}\textbf{50.7} & {\cellcolor[rgb]{0.925,0.957,1}}24.8 & {\cellcolor[rgb]{0.925,0.957,1}}\textbf{68.9} & {\cellcolor[rgb]{0.925,0.957,1}}51.5 & {\cellcolor[rgb]{0.925,0.957,1}}\textbf{61.9} & {\cellcolor[rgb]{0.925,0.957,1}}\textbf{49.6} & {\cellcolor[rgb]{0.925,0.957,1}}2,004 & {\cellcolor[rgb]{0.925,0.957,1}}5$\times$ & {\cellcolor[rgb]{0.925,0.957,1}}\textbf{34.6} & {\cellcolor[rgb]{0.925,0.957,1}}2,041 & {\cellcolor[rgb]{0.925,0.957,1}}5$\times$\\

    \bottomrule
    \end{tabular}

%% file: tabs/kv_cache.tex
\begin{tabular}
{l@{\hspace{0.05ex}}c@{\hspace{0.05ex}}c@{\hspace{0.05ex}}c@{\hspace{0.05ex}}c@{\hspace{0.05ex}}c@{\hspace{0.05ex}}c@{\hspace{0.05ex}}c@{\hspace{0.05ex}}c@{\hspace{0.05ex}}c@{\hspace{0.05ex}}c@{\hspace{0.05ex}}c@{\hspace{0.05ex}}c@{\hspace{0.05ex}}c@{\hspace{0.05ex}}c@{\hspace{0.05ex}}c@{\hspace{0.6ex}}c @{\hspace{0.6ex}} c }
    \toprule
\multirow{4}{*}{Method}& \multicolumn{3}{c}{Single-Document QA} & \multicolumn{3}{c}{Multi-Document QA}& \multicolumn{3}{c}{Summarization}& \multicolumn{3}{c}{Few-shot Learning} & \multicolumn{2}{c}{Synthetic} & \multicolumn{2}{c}{Code}
&\multirow{4}{*}{Average} \\
\cmidrule(lr){2-4}\cmidrule(lr){5-7}\cmidrule(lr){8-10}\cmidrule(lr){11-13}\cmidrule(lr){14-15}
& \rotatebox[origin=c]{40}{NrtvQA} & \rotatebox[origin=c]{40}{Qasper} & \rotatebox[origin=c]{40}{MF-en} & \rotatebox[origin=c]{40}{HotpotQA} & \rotatebox[origin=c]{40}{2WikiMQA} & \rotatebox[origin=c]{40}{Musique} & \rotatebox[origin=c]{40}{GovReport} & \rotatebox[origin=c]{40}{QMSum} & \rotatebox[origin=c]{40}{MultiNews} & \rotatebox[origin=c]{40}{TREC} & \rotatebox[origin=c]{40}{TriviaQA} & \rotatebox[origin=c]{40}{SAMSum} &  \rotatebox[origin=c]{40}{PCount} & \rotatebox[origin=c]{40}{~PRe~~} & \rotatebox[origin=c]{40}{RepoBench-P} & \rotatebox[origin=c]{40}{~LCC~~}  & \\
\toprule
\\
\multicolumn{16}{c}{\bf LLaMA 3.1 8B Instruct }\\
    \midrule
    All KV & {32.02} & {13.04} & {27.34} & {16.23} & {16.05} & {11.22} & { 34.52} & {23.41} & { 26.89} & { 73.0} & {91.64} &  {43.8} & {7.16} & {97.73} &49.25 &52.7 & { 38.32}  \\
   \midrule
     H2O-1024 &   21.98 & 10.96 & 23.17 & 16.06 & 15.16 & 10.15 & \bf 30.76 & 23.75 & \bf 26.32 & \bf 68.0 & 90.68 & 42.54 & 7.40 & 71.34 & \bf 51.6 & 46.4 & 34.77  \\ 
    SnapKV-1024  & {\bf 31.98} & {11.17} & {25.33} & {14.81} & {15.73} & {10.69} & {26.95} & {22.89} & {25.86} & {67.5} & {\bf 91.89} & {\bf 42.85} & {\bf 7.67} & {\bf 98.16} &48.7 &\bf 52.1 &{\bf 35.25} \\
    GemFilter-1024  
    & {20.71} & {11.00} & {\bf 29.28} & {19.12} & {\bf 17.01} & {13.01} & {30.37} & {21.75} & {25.17} & {63.0} & {90.70} & {42.50} & {7.15} & {92.22} & 35.0&38.5 &{34.50} \\
    \ehpc (\textit{ours})-1024  
    & {22.98} & {\bf 13.02} & { 27.41} & {\bf 20.64} & {16.97} & {\bf 13.99} & {29.49} &  {\bf 24.15} &{25.24} & {\bf 68.0} & {86.52} & {41.86} & {5.79} & {97.21} &42.8 & 48.9& {35.23} \\
   \midrule
    H2O-2048   & 23.16 & 11.68 & 25.09 & 16.17 & 15.22 & 9.93 & 32.13 & 23.32 & \bf 26.73 & 68.5 & 91.32 & \bf 43.97 & 6.52 & 72.79 & \bf 52.5 & 48.1 & 35.45 \\ 
    SnapKV-2048  & {\bf 31.45} & {11.94} & {26.24} & {15.73} & {16.03} & {11.66} & {29.64} & {23.24} & {26.44} & {69.5} & {91.48} & {42.68} & {7.21} & {98.03} &49.5 &\bf 52.6 & {35.80} \\
    GemFilter-2048  
    & {24.36} & {12.63} & {25.39} & {\bf 19.58} & {\bf 17.03} & {\bf 14.11} & {\bf 33.15} & {22.31} & {26.49} & {69.5} & {\bf 91.59} & {42.64} & {4.61} & {\bf 98.75} &38.8 &47.3 & {35.87} \\

    \ehpc (\textit{ours})-2048
    & {20.98} & {\bf 14.38} & {\bf 30.50} & {19.35} & {16.23} & {13.02} & {32.9} &  {\bf 24.54} &{26.72}  & {\bf 71.0} & {90.47} & {42.24} & {\bf 9.35} & {97.94} &44.9 &52.2 & {\bf 37.86} \\

\midrule
\\
\multicolumn{16}{c}{\bf Phi 3.5 Mini 3.8B Instruct}\\  \midrule
    All KV & {27.51} & {17.23} & {35.63} & {21.70} & {25.70} & {11.68} & {34.14} & {23.17} & {24.95} & {71.5} & {87.37} & {13.08} & {7.17} & {83.85} &46.0 &46.2 & {34.62} \\ 

   \midrule
    H2O-1024 &18.25 & 12.97 & 29.69 & 20.75 & 20.90 & 9.90 & \bf 32.02 & 21.69 & \bf 24.70 & 67.5 & 85.45 & 20.16 & 1.37 & 46.80 & \bf 46.6 & 43.3 & 31.38 \\
    SnapKV-1024 & {\bf 24.31} & {16.03} & {34.93} & {20.72} & {26.02} & {13.74} & {28.27} & {22.03} & {24.02} & {67.5} & {\bf 87.71} & {14.57} & {\bf 6.08} & {\bf 85.60} & 44.0&\bf 47.1 & {33.68} \\
    GemFilter-1024 
    & {16.57} & {18.29} & {35.91} & {24.22} & {26.10} & {9.70} & {30.29} & {18.96} & {23.64} & {64.5} & {85.85} & {23.02} & {0.20} & {81.12} &39.0 &40.7 &{32.74} \\
    \ehpc (\textit{ours})-1024 
    & {23.91} & {\bf 32.22} & {\bf 45.36} & {\bf 44.97} & {\bf 32.79} & {\bf 20.27} & {31.90} & {\bf 22.19} & {23.77} & {\bf 68.5} & {85.72} & {\bf 36.69} & {1.80} & {79.08} &38.7 & 41.6&{\bf 39.22} \\
   \midrule
   H2O-2048 & 18.26 & 14.05 & 32.4 & 20.03 & 22.51 & 10.30 & 32.86 & 21.43 & \bf 24.90 & 67.2 & 86.44 & 19.65 & 1.43 & 46.96 & \bf 47.89 & 44.71 & 31.94  \\
    SnapKV-2048  & {26.41} & {16.59} & {36.99} & {21.80} & {\bf 26.07} & {12.57} & {30.88} & {22.37} & {24.51} & {\bf 69.5} & {87.54} & {13.13} & {\bf 6.57} & {\bf 83.92} &45.30 &46.70 & {34.20} \\
    GemFilter-2048 
    & {19.63} & {14.84} & {35.99} & {21.38} & {19.72} & {10.13} & {32.39} & {21.24} & {24.71} & {65.0} & {86.49} & {20.47} & {2.17} & {69.50} &46.30 &\bf 48.10 &{31.69} \\
    \ehpc (\textit{ours})-2048 
    & {\bf 24.80} & {\bf 39.27} & {\bf 39.78} & {\bf 27.06} & {25.22} & {\bf 14.05} & {\bf 33.26} & {\bf 23.52} & {24.48} & {68.0} & {\bf 87.66} & {\bf 37.77} & {2.54} & {63.00} &47.31 &45.44 &{\bf 36.46} \\
\bottomrule
\end{tabular}